\documentstyle[aclap,epsf]{article}

\makeatletter
\newenvironment{algorithm}
  {\makeatletter
   \def\theenumi{\arabic{enumi}}
   
   \def\theenumii{\arabic{enumii}}
   
   \def\p@enumii{\theenumi.}

   \def\p@enumiii{\theenumi.\theenumii.}
   \makeatother
   \begin{enumerate}}%
  {\makeatletter
   
   \def\theenumi{\arabic{enumi}}
   
   \def\theenumii{\alph{enumii}}
   \def\p@enumii{\theenumi}

   \def\p@enumiii{\theenumi(\theenumii)}
   \makeatother
   \end{enumerate}}

\def\fileversion{v0.2}
\def\filedate{5/17/92}

\ifx\dialog@loaded\relax\endinput\else\let\dialog@loaded\relax\fi
\newcounter{diacount}
\newcounter{diacontinue}
\newcommand{\speaker}{A}
\newcommand{\listener}{B}
\newlength{\dialin}

\newlength{\myboxsize}
\newlength{\tmpboxsize}
\newlength{\subsize}
\newcounter{subcounter}[diacount]

\renewcommand{\thediacount}{\arabic{diacount}}

\newcommand{\edialog}
   {\end{list}}

\newcommand{\edialogat}
   {\end{list}}

\newcommand{\bdialog}[2]
  {\renewcommand{\speaker}{#1}
   \renewcommand{\listener}{#2}
   \settowidth{\dialin}{\speaker}
   \begin{list}
         {}
         {\usecounter{diacount}
          \setlength{\labelwidth}{7em}
          \settowidth{\myboxsize}{\speaker}
          \setlength{\rightmargin}{\leftmargin}
	  \addtolength{\leftmargin}{2em}}}

\newcommand{\bdialogcont}[2]
  {\renewcommand{\speaker}{#1}
   \renewcommand{\listener}{#2}
   \settowidth{\dialin}{\speaker}
   \setcounter{diacontinue}{\thediacount}
   \begin{list}
         {}
         {\usecounter{diacount}
	  \setcounter{diacount}{\thediacontinue}
          \setlength{\labelwidth}{7em}
          \settowidth{\myboxsize}{\speaker}
          \setlength{\rightmargin}{\leftmargin}
	  \addtolength{\leftmargin}{2em}}}

\newcommand{\speakerlab}
  {
   \refstepcounter{diacount}
   \settowidth{\tmpboxsize}{(\arabic{diacount}) }
   \addtolength{\tmpboxsize}{\myboxsize}
   \item[\hbox to \tmpboxsize{(\arabic{diacount}) \speaker}]}

\newcommand{\speakerlabsub}
  {\stepcounter{diacount}
   \refstepcounter{subcounter}
   \settowidth{\tmpboxsize}{(\arabic{diacount}\alph{subcounter})}
   \settowidth{\subsize}{\alph{subcounter}}
   \addtolength{\tmpboxsize}{\myboxsize}
   \addtolength{\tmpboxsize}{\subsize}
   \item[\hbox to \tmpboxsize{(\arabic{diacount}\alph{subcounter}) \speaker}]}

\newcommand{\speakerlabsubcont}
  {\refstepcounter{subcounter}
   \settowidth{\tmpboxsize}{(\arabic{diacount}\alph{subcounter})}
   \settowidth{\subsize}{\alph{subcounter}}
   \addtolength{\tmpboxsize}{\myboxsize}
   \addtolength{\tmpboxsize}{\subsize}
   \item[\hbox to \tmpboxsize{(\arabic{diacount}\alph{subcounter}) \speaker}]}

\newcommand{\speakerline}
  {
   \settowidth{\tmpboxsize}{(\arabic{diacount}) }
   \addtolength{\tmpboxsize}{\myboxsize}
   \item[\hbox to \tmpboxsize{\speaker}]}

\newcommand{\listenerlab}
  {
   \refstepcounter{diacount}
   \settowidth{\tmpboxsize}{(\arabic{diacount}) }
   \addtolength{\tmpboxsize}{\myboxsize}
   \item[\hbox to \tmpboxsize{(\arabic{diacount}) \listener}]}

\newcommand{\listenerline}
  {
   \settowidth{\tmpboxsize}{(\arabic{diacount}) }
   \addtolength{\tmpboxsize}{\myboxsize}
   \item[\hbox to \tmpboxsize{\listener}]}

\newcommand{\diacont}
  {\settowidth{\tmpboxsize}{(\arabic{diacount}) }
   \addtolength{\tmpboxsize}{\myboxsize}
   \item[\hbox to \tmpboxsize{}]}

\newcommand{\dialine}
  {
   \refstepcounter{diacount}
   \settowidth{\tmpboxsize}{(\arabic{diacount}) }
   \addtolength{\tmpboxsize}{\myboxsize}
   \item[\hbox to \tmpboxsize{(\arabic{diacount})}]}

\def\@listI{\leftmargin\leftmargini \parsep 0pt plus 1pt minus 1pt\topsep
2pt plus 1pt minus 1pt\itemsep 1pt plus 1pt minus 1pt}
\let\@listi\@listI
\@listi
\def\@listii{\leftmargin\leftmarginii
 \labelwidth\leftmarginii\advance\labelwidth-\labelsep
 \topsep 0pt plus 1pt minus 1pt
 \parsep 0pt plus 1pt minus 1pt
 \itemsep \parsep}
\def\@listiii{\leftmargin\leftmarginiii
 \labelwidth\leftmarginiii\advance\labelwidth-\labelsep
 \topsep 0pt plus 1pt minus 1pt
 \parsep \z@ \partopsep 1pt plus 0pt minus 1pt
 \itemsep \topsep}
\makeatother

\title{Response Generation in Collaborative Negotiation\thanks{This
material is based upon work supported by the National Science
Foundation under Grant No. IRI-9122026.}}

\author{Jennifer Chu-Carroll \and Sandra Carberry \\
        Department of Computer and Information Sciences \\
        University of Delaware \\
        Newark, DE 19716, USA \\
        E-mail: \{jchu,carberry\}@cis.udel.edu}

\date{}

\begin{document}

\maketitle

\begin{abstract}

In collaborative planning activities, since the agents are autonomous
and heterogeneous, it is inevitable that conflicts arise in their
beliefs during the planning process. In cases where such conflicts are
relevant to the task at hand, the agents should engage in {\em
collaborative negotiation} as an attempt to square away the
discrepancies in their beliefs. This paper presents a computational
strategy for detecting conflicts regarding proposed beliefs and for
engaging in collaborative negotiation to resolve the conflicts that
warrant resolution. Our model is capable of selecting the most
effective aspect to address in its pursuit of conflict resolution in
cases where multiple conflicts arise, and of selecting appropriate
evidence to justify the need for such modification. Furthermore, by
capturing the negotiation process in a recursive {\em
Propose-Evaluate-Modify} cycle of actions, our model can successfully
handle embedded negotiation subdialogues.

\end{abstract}

\section{Introduction}

In collaborative consultation dialogues, the consultant and the
executing agent collaborate on developing a plan to achieve the
executing agent's domain goal. Since agents are autonomous and
heterogeneous, it is inevitable that conflicts in their beliefs arise
during the planning process. In such cases, collaborative agents
should attempt to {\em square away} \cite{jos_mk82} the conflicts by
engaging in collaborative negotiation to determine what should
constitute their shared plan of actions and shared beliefs.
Collaborative negotiation differs from non-collaborative negotiation
and argumentation mainly in the {\em attitude} of the participants,
since collaborative agents are not self-centered, but act in a way as
to benefit the agents as a group. Thus, when facing a conflict, a
collaborative agent should not automatically reject a belief with
which she does not agree; instead, she should evaluate the belief and
the evidence provided to her and adopt the belief if the evidence is
convincing.  On the other hand, if the evaluation indicates that the
agent should maintain her original belief, she should attempt to
provide sufficient justification to convince the other agent to adopt
this belief if the belief is relevant to the task at hand.

This paper presents a model for engaging in collaborative negotiation
to resolve conflicts in agents' beliefs about domain knowledge. Our
model 1) detects conflicts in beliefs and initiates a negotiation
subdialogue only when the conflict is relevant to the current task, 2)
selects the most effective aspect to address in its pursuit of
conflict resolution when multiple conflicts exist, 3) selects
appropriate evidence to justify the system's proposed modification of
the user's beliefs, and 4) captures the negotiation process in a
recursive {\em Propose-Evaluate-Modify} cycle of actions, thus
enabling the system to handle embedded negotiation subdialogues.

\section{Related Work}

Researchers have studied the analysis and generation of arguments
\cite{biretal_aaai80,rei_ijcai81,coh_cl87,syc_ijcai89,qui_coling92,may_aaai93};
however, agents engaging in argumentative dialogues are solely
interested in winning an argument and thus exhibit different behavior
from collaborative agents. Sidner \shortcite{sid_aaaiws92,sid_aaai94}
formulated an artificial language for modeling collaborative discourse
using proposal/acceptance and proposal/rejection sequences; however,
her work is descriptive and does not specify response generation
strategies for agents involved in collaborative interactions.

Webber and Joshi \shortcite{web_jos_coling82} have noted the
importance of a cooperative system providing support for its
responses. They identified strategies that a system can adopt in
justifying its beliefs; however, they did not specify the criteria
under which each of these strategies should be selected. Walker
\shortcite{wal_coling94} described a method of determining when to
include optional warrants to justify a claim based on factors such as
communication cost, inference cost, and cost of memory
retrieval. However, her model focuses on determining when to include
informationally redundant utterances, whereas our model determines
whether or not justification is needed for a claim to be convincing
and, if so, selects appropriate evidence from the system's private
beliefs to support the claim.

Caswey et al. \cite{cawetal_aisb93,logetal_tr94} introduced the idea
of utilizing a belief revision mechanism \cite{gal_br92} to predict
whether a set of evidence is sufficient to change a user's existing
belief and to generate responses for information retrieval dialogues
in a library domain. They argued that in the library dialogues they
analyzed, ``in no cases does negotiation extend beyond the initial
belief conflict and its immediate resolution.'' \cite[page
141]{logetal_tr94}.  However, our analysis of naturally-occurring
consultation dialogues \cite{coltrans85,sri92} shows that in other
domains conflict resolution does extend beyond a single exchange of
conflicting beliefs; therefore we employ a recursive model for
collaboration that captures extended negotiation and represents the
structure of the discourse.  Furthermore, their system deals with a
single conflict, while our model selects a focus in its pursuit of
conflict resolution when multiple conflicts arise. In addition, we
provide a process for selecting among multiple possible pieces of
evidence.

\section{Features of Collaborative Negotiation}

Collaborative negotiation occurs when conflicts arise among agents
developing a shared plan\footnote{The notion of {\em shared plan} has
been used in \cite{gro_sid_ic90,all_snlw91}.} during collaborative
planning. A collaborative agent is driven by the goal of developing a
plan that best satisfies the interests of all the agents as a group,
instead of one that maximizes his own interest. This results in
several distinctive features of collaborative negotiation: 1) A
collaborative agent does not insist on winning an argument, and may
change his beliefs if another agent presents convincing justification
for an opposing belief. This differentiates collaborative negotiation
from argumentation
\cite{biretal_aaai80,rei_ijcai81,coh_cl87,qui_coling92}. 2) Agents
involved in collaborative negotiation are open and honest with one
another; they will not deliberately present false information to other
agents, present information in such a way as to mislead the other
agents, or strategically hold back information from other agents for
later use. This distinguishes collaborative negotiation from
non-collaborative negotiation such as labor negotiation
\cite{syc_ijcai89}. 3) Collaborative agents are interested in others'
beliefs in order to decide whether to revise their own beliefs so as
to come to agreement \cite{chu_car_ijcai95}. Although agents involved
in argumentation and non-collaborative negotiation take other agents'
beliefs into consideration, they do so mainly to find weak points in
their opponents' beliefs and attack them to win the argument.

In our earlier work, we built on Sidner's proposal/acceptance and
proposal/rejection sequences \cite{sid_aaai94} and developed a model
that captures collaborative planning processes in a {\em
Propose-Evaluate-Modify} cycle of actions \cite{chu_car_aaai94}. This
model views collaborative planning as agent A {\em proposing} a set of
actions and beliefs to be incorporated into the shared plan being
developed, agent B {\em evaluating} the proposal to determine whether
or not he accepts the proposal and, if not, agent B proposing a set of
{\em modifications} to A's original proposal. The proposed
modifications will again be evaluated by A, and if conflicts arise,
she may propose modifications to B's previously proposed
modifications, resulting in a recursive process.  However, our
research did not specify, in cases where multiple conflicts arise, how
an agent should identify which part of an unaccepted proposal to
address or how to select evidence to support the proposed
modification. This paper extends that work by incorporating into the
modification process a strategy to determine the aspect of the
proposal that the agent will address in her pursuit of conflict
resolution, as well as a means of selecting appropriate evidence to
justify the need for such modification.

\section{Response Generation in Collaborative Negotiation}

In order to capture the agents' intentions conveyed by their
utterances, our model of collaborative negotiation utilizes an
enhanced version of the dialogue model described in
\cite{lam_car_acl91} to represent the current status of the
interaction. The enhanced dialogue model has four levels: the {\em
domain} level which consists of the domain plan being constructed for
the user's later execution, the {\em problem-solving} level which
contains the actions being performed to construct the domain plan, the
{\em belief} level which consists of the mutual beliefs pursued during
the planning process in order to further the problem-solving
intentions, and the {\em discourse} level which contains the
communicative actions initiated to achieve the mutual beliefs
\cite{chu_car_aaai94}. This paper focuses on the evaluation and
modification of proposed {\em beliefs}, and details a strategy for
engaging in collaborative negotiations.

\subsection{Evaluating Proposed Beliefs}
\label{evaluate}

Our system maintains a set of beliefs about the domain and about the
user's beliefs. Associated with each belief is a {\em strength} that
represents the agent's confidence in holding that belief. We model the
strength of a belief using {\em endorsements}, which are explicit
records of factors that affect one's certainty in a hypothesis
\cite{coh_book85}, following \cite{gal_br92,logetal_tr94}. Our
endorsements are based on the semantics of the utterance used to
convey a belief, the level of expertise of the agent conveying the
belief, stereotypical knowledge, etc.

The belief level of the dialogue model consists of mutual beliefs
proposed by the agents' discourse actions. When an agent proposes a
new belief and gives (optional) supporting evidence for it, this set
of proposed beliefs is represented as a belief tree, where the belief
represented by a child node is intended to support that represented by
its parent. The root nodes of these belief trees (top-level beliefs)
contribute to problem-solving actions and thus affect the domain plan
being developed. Given a set of newly proposed beliefs, the system
must decide whether to accept the proposal or to initiate a
negotiation dialogue to resolve conflicts.  The evaluation of proposed
beliefs starts at the leaf nodes of the proposed belief trees since
acceptance of a piece of proposed evidence may affect acceptance of
the parent belief it is intended to support. The process continues
until the top-level proposed beliefs are evaluated. Conflict
resolution strategies are invoked only if the top-level proposed
beliefs are not accepted because if collaborative agents agree on a
belief relevant to the domain plan being constructed, it is irrelevant
whether they agree on the evidence for that belief
\cite{youetal_cogsci94}.

In determining whether to accept a proposed belief or evidential
relationship, the evaluator first constructs an evidence set
containing the system's evidence that supports or attacks \_bel and
the evidence accepted by the system that was proposed by the user as
support for \_bel. Each piece of evidence contains a belief {\em
\_bel$_i$}, and an evidential relationship {\em
supports(\_bel$_i$,\_bel)}. Following Walker's {\em weakest link
assumption} \cite{wal_coling92} the strength of the evidence is the
weaker of the strength of the belief and the strength of the
evidential relationship. The evaluator then employs a simplified
version of Galliers' belief revision mechanism\footnote{For details on
how our model determines the acceptance of a belief using the ranking
of endorsements proposed by Galliers, see \cite{chu_phd}.}
\cite{gal_br92,logetal_tr94} to compare the strengths of the evidence
that supports and attacks \_bel. If the strength of one set of
evidence strongly outweighs that of the other, the decision to accept
or reject \_bel is easily made. However, if the difference in their
strengths does not exceed a pre-determined threshold, the evaluator
has insufficient information to determine whether to adopt \_bel and
therefore will initiate an {\em information-sharing subdialogue}
\cite{chu_car_ijcai95} to share information with the user so that each
of them can knowledgably re-evaluate the user's original proposal. If,
during information-sharing, the user provides convincing support for a
belief whose negation is held by the system, the system may adopt the
belief after the re-evaluation process, thus resolving the conflict
without negotiation.

\subsubsection{Example}
\label{example}

To illustrate the evaluation of proposed beliefs, consider the
following utterances:
\bdialog{S:}{U:}
\em

\speakerlab \label{ai} I think Dr.~Smith is teaching AI next semester.

\listenerlab \label{notai} Dr.~Smith is not teaching AI.

\dialine \label{sabbatical} He is going on sabbatical next year.

\edialog
\begin{figure}
\centerline{\epsfysize=2.1in\epsffile{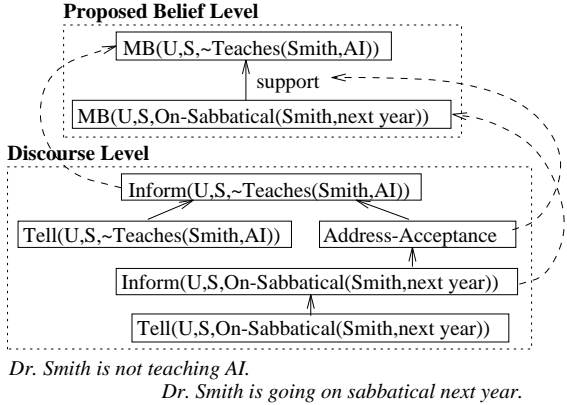}}
\caption{Belief and Discourse Levels for (\protect{\ref{notai}}) and
(\protect{\ref{sabbatical}})}
\label{teach}
\vspace{1ex}
\hrule
\end{figure}

\noindent Figure~\ref{teach} shows the belief and discourse levels of
the dialogue model that captures utterances (\ref{notai}) and
(\ref{sabbatical}). The belief evaluation process will start with the
belief at the leaf node of the proposed belief tree, {\em
On-Sabbatical(Smith,next year))}. The system will first gather its
evidence pertaining to the belief, which includes 1) a warranted
belief\footnote{The strength of a belief is classified as: {\em
warranted}, {\em strong}, or {\em weak}, based on the endorsement of
the belief.} that Dr.~Smith has postponed his sabbatical until 1997
({\em Postponed-Sabbatical(Smith,1997)}), 2) a warranted belief that
Dr.~Smith postponing his sabbatical until 1997 supports the belief
that he is not going on sabbatical next year ({\em
supports(Postponed-Sabbatical(Smith,1997),
$\lnot$On-Sabbatical(Smith,next year))}, 3) a strong belief that
Dr.~Smith will not be a visitor at IBM next year ({\em
$\lnot$visitor(Smith, IBM, next year)}), and 4) a warranted belief
that Dr.~Smith not being a visitor at IBM next year supports the
belief that he is not going on sabbatical next year ({\em
supports($\lnot$visitor(Smith, IBM, next year),
$\lnot$On-Sabbatical(Smith, next year))}, perhaps because Dr.~Smith
has expressed his desire to spend his sabbatical only at IBM). The
belief revision mechanism will then be invoked to determine the
system's belief about {\em On-Sabbatical(Smith, next year)} based on
the system's own evidence and the user's statement. Since beliefs (1)
and (2) above constitute a warranted piece of evidence against the
proposed belief and beliefs (3) and (4) constitute a strong piece of
evidence against it, the system will not accept {\em
On-Sabbatical(Smith, next year)}.

The system believes that being on sabbatical implies a faculty member
is not teaching any courses; thus the proposed evidential relationship
will be accepted.  However, the system will not accept the top-level
proposed belief, {\em $\lnot$Teaches(Smith, AI)}, since the system has
a prior belief to the contrary (as expressed in utterance (1)) and the
only evidence provided by the user was an implication whose antecedent
was not accepted.

\subsection{Modifying Unaccepted Proposals}

The {\em collaborative planning principle} in
\cite{whi_ste_acl88,wal_coling92} suggests that ``conversants must
provide evidence of a detected discrepancy in belief as soon as
possible.'' Thus, once an agent detects a relevant conflict, she must
notify the other agent of the conflict and initiate a negotiation
subdialogue to resolve it --- to do otherwise is to fail in her
responsibility as a collaborative agent. We capture the attempt to
resolve a conflict with the problem-solving action {\em
Modify-Proposal}, whose goal is to modify the proposal to a form that
will potentially be accepted by both agents. When applied to belief
modification, {\em Modify-Proposal} has two specializations: {\em
Correct-Node}, for when a proposed belief is not accepted, and {\em
Correct-Relation}, for when a proposed evidential relationship is not
accepted. Figure~\ref{recipes} shows the problem-solving
recipes\footnote{A recipe \cite{pol_acl86} is a template for
performing actions. It contains the applicability conditions for
performing an action, the subactions comprising the body of an action,
etc.} for {\em Correct-Node} and its subaction, {\em Modify-Node},
that is responsible for the actual modification of the proposal. The
applicability conditions\footnote{Applicability conditions are
conditions that must already be satisfied in order for an action to be
reasonable to pursue, whereas an agent can try to achieve unsatisfied
preconditions.}  of {\em Correct-Node} specify that the action can
only be invoked when \_s1 believes that \_node is not acceptable while
\_s2 believes that it is (when \_s1 and \_s2 disagree about the
proposed belief represented by \_node). However, since this is a
collaborative interaction, the actual modification can only be
performed when both \_s1 and \_s2 believe that \_node is not
acceptable --- that is, the conflict between \_s1 and \_s2 must have
been resolved. This is captured by the applicability condition and
precondition of {\em Modify-Node}. The attempt to satisfy the
precondition causes the system to post as a mutual belief to be
achieved the belief that \_node is not acceptable, leading the system
to adopt discourse actions to change \_s2's beliefs, thus initiating a
collaborative negotiation subdialogue.\footnote{This subdialogue is
considered an {\em interrupt} by Whittaker, Stenton, and Walker
\cite{whi_ste_acl88,wal_whi_acl90}, initiated to negotiate the truth
of a piece of information.  However, the utterances they classify as
{\em interrupts} include not only our negotiation subdialogues,
generated for the purpose of {\em modifying} a proposal, but also
clarification subdialogues, and {\em information-sharing} subdialogues
\cite{chu_car_ijcai95}, which we contend should be part of the {\em
evaluation} process.}

\begin{figure}
\footnotesize
\begin{tabbing}
Appl Cond: \=  \kill
Action: \> Correct-Node(\_s1, \_s2, \_proposed)\\
Type: \> Decomposition \\
Appl Cond: \> believe(\_s1,$\lnot$acceptable(\_node)) \\
\> believe(\_s2, acceptable(\_node))\\
Const: \> error-in-plan(\_node,\_proposed) \\
Body: \> Modify-Node(\_s1,\_s2,\_proposed,\_node) \\
      \> Insert-Correction(\_s1, \_s2, \_proposed)\\
Goal: \> acceptable(\_proposed) \\ \\
Action: \> Modify-Node(\_s1,\_s2,\_proposed,\_node) \\
Type: \> Specialization \\
Appl Cond: \> believe(\_s1,$\lnot$acceptable(\_node)) \\
Precond: \> believe(\_s2,$\lnot$acceptable(\_node)) \\
Body: \> Remove-Node(\_s1,\_s2,\_proposed,\_node) \\
      \> Alter-Node(\_s1,\_s2,\_proposed,\_node) \\
Goal: \> modified(\_proposed)
\end{tabbing}
\caption{The {\em Correct-Node} and {\em Modify-Node} Recipes}
\label{recipes}
\vspace{1ex}
\hrule
\end{figure}

\subsubsection{Selecting the Focus of Modification}

When multiple conflicts arise between the system and the user
regarding the user's proposal, the system must identify the aspect of
the proposal on which it should focus in its pursuit of conflict
resolution. For example, in the case where {\em Correct-Node} is
selected as the specialization of {\em Modify-Proposal}, the system
must determine how the parameter \_node in {\em Correct-Node} should
be instantiated. The goal of the modification process is to resolve
the agents' conflicts regarding the unaccepted top-level proposed
beliefs. For each such belief, the system could provide evidence
against the belief itself, address the unaccepted evidence proposed by
the user to eliminate the user's justification for the belief, or
both. Since collaborative agents are expected to engage in effective
and efficient dialogues, the system should address the unaccepted
belief that it predicts will most quickly resolve the top-level
conflict. Therefore, for each unaccepted top-level belief, our process
for selecting the focus of modification involves two steps:
identifying a candidate foci tree from the proposed belief tree, and
selecting a focus from the candidate foci tree using the heuristic
``attack the belief(s) that will most likely resolve the conflict
about the top-level belief.''  A candidate foci tree contains the
pieces of evidence in a proposed belief tree which, if disbelieved by
the user, might change the user's view of the unaccepted top-level
proposed belief (the root node of that belief tree). It is identified
by performing a depth-first search on the proposed belief tree. When a
node is visited, both the belief and the evidential relationship
between it and its parent are examined. If both the belief and
relationship were accepted by the evaluator, the search on the current
branch will terminate, since once the system accepts a belief, it is
irrelevant whether it accepts the user's support for that belief
\cite{youetal_cogsci94}. Otherwise, this piece of evidence will be
included in the candidate foci tree and the system will continue to
search through the evidence in the belief tree proposed as support for
the unaccepted belief and/or evidential relationship.

\begin{figure}
\footnotesize

{\bf Select-Focus-Modification}(\_bel):
\begin{algorithm}

\item \_bel.u-evid $\leftarrow$ system's beliefs about the user's
evidence pertaining to \_bel

      \_bel.s-attack $\leftarrow$ system's own evidence against \_bel

\item \label{leaf} If \_bel is a leaf node in the candidate foci tree,

   \begin{algorithm}

   \item If {\bf Predict}(\_bel, \_bel.u-evid + \_bel.s-attack) =
$\lnot$\_bel

         then \_bel.focus $\leftarrow$ \_bel; return

   \item Else \_bel.focus $\leftarrow$ nil; return

   \end{algorithm}

\item \label{recurse} Select focus for each of \_bel's children
in the candidate foci tree, \_bel$_1$,$\ldots$,\_bel$_n$:

   \begin{algorithm}

   \item If supports(\_bel$_i$,\_bel) is accepted but
\_bel$_i$ is not, {\bf Select-Focus-Modification}(\_bel$_i$).

   \item Else if \_bel$_i$ is accepted but supports(\_bel$_i$,\_bel)
is not, {\bf Select-Focus-Modification}(\_bel$_i$,\_bel).

   \item Else {\bf Select-Focus-Modification}(\_bel$_i$) and {\bf
Select-Focus-Modification}(supports(\_bel$_i$,\_bel))

   \end{algorithm}

\item \label{predict} Choose between attacking the proposed evidence
for \_bel and attacking \_bel itself:

   \begin{algorithm}

   \item cand-set $\leftarrow$ \{\_bel$_i$ | \_bel$_i$ $\in$
unaccepted user evidence for \_bel $\land$ \_bel$_i$.focus $\neq$
nil\}

   \item \label{children} {\em // Check if addressing \_bel's
unaccepted evidence is sufficient}

If {\bf Predict}(\_bel, \_bel.u-evid - cand-set) =
$\lnot\_bel$ (i.e., the user's disbelief in all unaccepted evidence
which the system can refute will cause him to reject \_bel),

   min-set $\leftarrow$ {\bf Select-Min-Set}(\_bel,cand-set)

   \_bel.focus $\leftarrow \bigcup_{\mbox{\_bel}_i \in
\mbox{\_min-set}}$ \_bel$_i$.focus

   \item \label{bel} {\em // Check if addressing \_bel itself is
sufficient}

Else if {\bf Predict}(\_bel, \_bel.u-evid + \_bel.s-attack) =
$\lnot\_bel$ (i.e., the system's evidence against \_bel will cause the
user to reject \_bel),

   \_bel.focus $\leftarrow$ \_bel

   \item \label{both} {\em // Check if addressing both \_bel and its
unaccepted evidence is sufficient}

Else if {\bf Predict}(\_bel, \_bel.s-attack + \_bel.u-evid -
cand-set) = $\lnot\_bel$,

   min-set $\leftarrow$ {\bf Select-Min-Set}(\_bel, cand-set + \_bel)

   \_bel.focus $\leftarrow \bigcup_{\mbox{\_bel}_i \in
\mbox{\_min-set}}$ \_bel$_i$.focus $\cup$ \_bel

   \item \label{nil} Else \_bel.focus $\leftarrow$ nil

   \end{algorithm}

\end{algorithm}

\caption{Selecting the Focus of Modification}
\label{algorithm}
\vspace{1ex}
\hrule
\end{figure}

Once a candidate foci tree is identified, the system should select the
focus of modification based on the likelihood of each choice changing
the user's belief about the top-level belief. Figure~\ref{algorithm}
shows our algorithm for this selection process. Given an unaccepted
belief (\_bel) and the beliefs proposed to support it, {\bf
Select-Focus-Modification} will annotate \_bel with 1) its focus of
modification (\_bel.focus), which contains a set of beliefs (\_bel
and/or its descendents) which, if disbelieved by the user, are
predicted to cause him to disbelieve \_bel, and 2) the system's
evidence against \_bel itself (\_bel.s-attack).

{\bf Select-Focus-Modification} determines whether to attack \_bel's
supporting evidence separately, thereby eliminating the user's reasons
for holding \_bel, to attack \_bel itself, or both.  However, in
evaluating the effectiveness of attacking the proposed evidence for
\_bel, the system must determine whether or not it is possible to
successfully refute a piece of evidence (i.e., whether or not the
system believes that sufficient evidence is available to convince the
user that a piece of proposed evidence is invalid), and if so, whether
it is more effective to attack the evidence itself or its
support. Thus the algorithm recursively applies itself to the evidence
proposed as support for \_bel which was not accepted by the system
(step~\ref{recurse}). In this recursive process, the algorithm
annotates each unaccepted belief or evidential relationship proposed
to support \_bel with its focus of modification (\_bel$_i$.focus) and
the system's evidence against it (\_bel$_i$.s-attack). \_bel$_i$.focus
contains the beliefs selected to be addressed in order to change the
user's belief about \_bel$_i$, and its value will be nil if the system
predicts that insufficient evidence is available to change the user's
belief about \_bel$_i$.

Based on the information obtained in step~\ref{recurse}, {\bf
Select-Focus-Modification} decides whether to attack the evidence
proposed to support \_bel, or \_bel itself (step~\ref{predict}). Its
preference is to address the unaccepted evidence, because McKeown's
focusing rules suggest that continuing a newly introduced topic (about
which there is more to be said) is preferable to returning to a
previous topic \cite{mck_book85}. Thus the algorithm first considers
whether or not attacking the user's support for \_bel is sufficient to
convince him of $\lnot$\_bel (step~\ref{children}). It does so by
gathering (in {\em cand-set}) evidence proposed by the user as direct
support for \_bel but which was not accepted by the system and which
the system predicts it can successfully refute (i.e., \_bel$_i$.focus
is not nil). The algorithm then hypothesizes that the user has changed
his mind about each belief in {\em cand-set} and predicts how this
will affect the user's belief about \_bel (step~\ref{children}). If
the user is predicted to accept $\lnot$\_bel under this hypothesis,
the algorithm invokes {\bf Select-Min-Set} to select a minimum subset
of {\em cand-set} as the unaccepted beliefs that it would actually
pursue, and the focus of modification (\_bel.focus) will be the union
of the focus for each of the beliefs in this minimum subset.

If attacking the evidence for \_bel does not appear to be sufficient
to convince the user of $\lnot$\_bel, the algorithm checks whether
directly attacking \_bel will accomplish this goal. If providing
evidence directly against \_bel is predicted to be successful, then
the focus of modification is \_bel itself (step~\ref{bel}). If
directly attacking \_bel is also predicted to fail, the algorithm
considers the effect of attacking both \_bel and its unaccepted
proposed evidence by combining the previous two prediction processes
(step~\ref{both}). If the combined evidence is still predicted to
fail, the system does not have sufficient evidence to change the
user's view of \_bel; thus, the focus of modification for \_bel is nil
(step~\ref{nil}).\footnote{In collaborative dialogues, an agent should
reject a proposal only if she has strong evidence against it. When an
agent does not have sufficient information to determine the acceptance
of a proposal, she should initiate an {\em information-sharing
subdialogue} to share information with the other agent and re-evaluate
the proposal \cite{chu_car_ijcai95}. Thus, further research is needed
to determine whether or not the focus of modification for a rejected
belief will ever be nil in collaborative dialogues.} Notice that
steps~\ref{leaf} and \ref{predict} of the algorithm invoke a function,
{\bf Predict}, that makes use of the belief revision mechanism
\cite{gal_br92} discussed in Section~\ref{evaluate} to predict the
user's acceptance or unacceptance of \_bel based on the system's
knowledge of the user's beliefs and the evidence that could be
presented to him \cite{logetal_tr94}. The result of {\bf
Select-Focus-Modification} is a set of user beliefs (in \_bel.focus)
that need to be modified in order to change the user's belief about
the unaccepted top-level belief.  Thus, the negations of these beliefs
will be posted by the system as mutual beliefs to be achieved in order
to perform the {\em Modify} actions.

\subsubsection{Selecting Justification for a Claim}

Studies in communication and social psychology have shown that
evidence improves the persuasiveness of a message
\cite{luc_mcc_sscj78,rey_bur_cy83,pet_cac_jpsp84,ham_wjsc85}.
Research on the quantity of evidence indicates that there is no
optimal amount of evidence, but that the use of high-quality evidence
is consistent with persuasive effects \cite{rei_hcr88}. On the other
hand, Grice's maxim of quantity \cite{gri_ss75} specifies that one
should not contribute more information than is
required.\footnote{Walker \shortcite{wal_coling94} has shown the
importance of IRU's (Informationally Redundant Utterances) in
efficient discourse. We leave including appropriate IRU's for future
work.} Thus, it is important that a collaborative agent selects
sufficient and effective, but not excessive, evidence to justify an
intended mutual belief.

To convince the user of a belief, \_bel, our system selects
appropriate justification by identifying beliefs that could be used to
support \_bel and applying filtering heuristics to them. The system
must first determine whether justification for \_bel is needed by
predicting whether or not merely informing the user of \_bel will be
sufficient to convince him of \_bel. If so, no justification will be
presented. If justification is predicted to be necessary, the system
will first construct the justification chains that could be used to
support \_bel. For each piece of evidence that could be used to
directly support \_bel, the system first predicts whether the user
will accept the evidence without justification. If the user is
predicted not to accept a piece of evidence (evid$_i$), the system
will augment the evidence to be presented to the user by posting
evid$_i$ as a mutual belief to be achieved, and selecting propositions
that could serve as justification for it. This results in a recursive
process that returns a chain of belief justifications that could be
used to support \_bel.

Once a set of beliefs forming justification chains is identified, the
system must then select from this set those belief chains which, when
presented to the user, are predicted to convince the user of
\_bel. Our system will first construct a singleton set for each such
justification chain and select the sets containing justification
which, when presented, is predicted to convince the user of \_bel.  If
no single justification chain is predicted to be sufficient to change
the user's beliefs, new sets will be constructed by combining the
single justification chains, and the selection process is repeated.
This will produce a set of possible candidate justification chains,
and three heuristics will then be applied to select from among
them. The first heuristic prefers evidence in which the system is most
confident since high-quality evidence produces more attitude change
than any other evidence form \cite{luc_mcc_sscj78}.  Furthermore, the
system can better justify a belief in which it has high confidence
should the user not accept it. The second heuristic prefers evidence
that is novel to the user, since studies have shown that evidence is
most persuasive if it is previously unknown to the hearer
\cite{wye_jesp70,mor_cm87}.  The third heuristic is based on Grice's
maxim of quantity and prefers justification chains that contain the
fewest beliefs.

\subsubsection{Example}

After the evaluation of the dialogue model in Figure~\ref{teach}, {\em
Modify-Proposal} is invoked because the top-level proposed belief is
not accepted.  In selecting the focus of modification, the system will
first identify the candidate foci tree and then invoke the {\bf
Select-Focus-Modification} algorithm on the belief at the root node of
the candidate foci tree. The candidate foci tree will be identical to
the proposed belief tree in Figure~\ref{teach} since both the
top-level proposed belief and its proposed evidence were rejected
during the evaluation process. This indicates that the focus of
modification could be either {\em $\lnot$Teaches(Smith,AI)} or {\em
On-Sabbatical(Smith, next year)} (since the evidential relationship
between them was accepted).  When {\bf Select-Focus-Modification} is
applied to {\em $\lnot$Teaches(Smith,AI)}, the algorithm will first be
recursively invoked on {\em On-Sabbatical(Smith, next year)} to
determine the focus for modifying the child belief (step 3.1 in
Figure~\ref{algorithm}). Since the system has two pieces of evidence
against {\em On-Sabbatical(Smith, next year)}, 1) a warranted piece of
evidence containing {\em Postponed-Sabbatical(Smith,1997)} and {\em
supports(Postponed-Sabbatical(Smith,1997),$\lnot$On-Sabbatical(Smith,
next year))}, and 2) a strong piece of evidence containing {\em
$\lnot$visitor(Smith,IBM,next year)} and {\em
supports($\lnot$visitor(Smith,IBM,next
year),$\lnot$On-Sabbatical(Smith,next year))}, the evidence is
predicted to be sufficient to change the user's belief in {\em
On-Sabbatical(Smith,next year)}, and hence {\em
$\lnot$Teaches(Smith,AI)}; thus, the focus of modification will be
{\em On-Sabbatical(Smith,next year)}. The {\em Correct-Node}
specialization of {\em Modify-Proposal} will be invoked since the
focus of modification is a belief, and in order to satisfy the
precondition of {\em Modify-Node} (Figure~\ref{recipes}), {\em MB(S,U,
$\lnot$On-Sabbatical(Smith,next year))} will be posted as a mutual
belief to be achieved.

Since the user has a warranted belief in {\em On-Sabbatical(Smith,next
year)} (indicated by the semantic form of utterance
(\ref{sabbatical})), the system will predict that merely informing the
user of the intended mutual belief is not sufficient to change his
belief; therefore it will select justification from the two available
pieces of evidence supporting {\em $\lnot$On-Sabbatical(Smith,next
year)} presented earlier.  The system will predict that either piece
of evidence combined with the proposed mutual belief is sufficient to
change the user's belief; thus, the filtering heuristics are applied.
The first heuristic will cause the system to select {\em
Postponed-Sabbatical(Smith, 1997)} and {\em
supports(Postponed-Sabbatical(Smith, 1997),$\lnot$On-Sabbatical(Smith,
next year))} as support, since it is the evidence in which the system
is more confident.

The system will try to establish the mutual beliefs\footnote{Only {\em
MB(S,U,Postponed-Sabbatical(Smith, 1997))} will be proposed as
justification because the system believes that the evidential
relationship needed to complete the inference is held by a
stereotypical user.} as an attempt to satisfy the precondition of {\em
Modify-Node}. This will cause the system to invoke {\em Inform}
discourse actions to generate the following utterances:

\bdialogcont{S:}{}
\em

\speakerlab \label{notsab} Dr.~Smith is not going on sabbatical next
year.

\dialine \label{postpone} He postponed his sabbatical until 1997.

\edialog

\noindent If the user accepts the system's utterances, thus satisfying
the precondition that the conflict be resolved, {\em Modify-Node} can
be performed and changes made to the original proposed beliefs.
Otherwise, the user may propose modifications to the system's proposed
modifications, resulting in an embedded negotiation subdialogue.

\section{Conclusion}

This paper has presented a computational strategy for engaging in
collaborative negotiation to square away conflicts in agents'
beliefs. The model captures features specific to collaborative
negotiation. It also supports effective and efficient dialogues by
identifying the focus of modification based on its predicted success
in resolving the conflict about the top-level belief and by using
heuristics motivated by research in social psychology to select a set
of evidence to justify the proposed modification of beliefs.
Furthermore, by capturing collaborative negotiation in a cycle of {\em
Propose-Evaluate-Modify} actions, the evaluation and modification
processes can be applied recursively to capture embedded negotiation
subdialogues.

\section*{Acknowledgments}

Discussions with Candy Sidner, Stephanie Elzer, and Kathy McCoy have
been very helpful in the development of this work. Comments from the
anonymous reviewers have also been very useful in preparing the final
version of this paper.

\bibliographystyle{acl}

\end{document}